# Further Results on "Velocity/Position Integration Formula (I): Application to In-Flight Coarse Alignment"

Yuanxin Wu

**Abstract—This note improves our above-mentioned recent work [1] by effectively depressing the adverse effect of the lever arm on attitude estimation.**

**Index Terms— attitude alignment, lever arm effect**

Our recent work [1] proposed a coarse in-motion alignment method, in which the adverse effect of the Global Positioning System (GPS) antenna lever arm was reported. The current note improves [1] by accounting for the prior knowledge of the lever arm.

This note uses the same set of symbols as in [1] for brevity. Suppose the GPS antenna is rigidly fixed relative to the inertial navigation system (INS) and the lever arm from INS to the GPS antenna is $\mathbf{l}^b$, expressed in the body frame, then the GPS antenna velocity and position is related to the INS position and velocity by [2]

$$\mathbf{p}_{gps} \approx \mathbf{p} + \mathbf{R}_c \mathbf{C}_b^n \mathbf{l}^b \tag{1}$$

$$\mathbf{v}_{gps}^n = \mathbf{v}^n + \mathbf{C}_b^n \left( \boldsymbol{\omega}_{eb}^b \times \mathbf{l}^b \right). \tag{2}$$

where the symbols are well defined in [1]. Substituting $\mathbf{v}^n = \mathbf{v}_{gps}^n - \mathbf{C}_b^n \left( \boldsymbol{\omega}_{eb}^b \times \mathbf{l}^b \right)$ into (2) in [1]

$$\dot{\mathbf{v}}_{gps}^n - \mathbf{C}_b^n \boldsymbol{\omega}_{nb}^b \times \left( \boldsymbol{\omega}_{eb}^b \times \mathbf{l}^b \right) - \mathbf{C}_b^n \left( \dot{\boldsymbol{\omega}}_{eb}^b \times \mathbf{l}^b \right) = \mathbf{C}_b^n \mathbf{f}^b - \left( 2\boldsymbol{\omega}_{ie}^n + \boldsymbol{\omega}_{en}^n \right) \times \left( \mathbf{v}_{gps}^n - \mathbf{C}_b^n \left( \boldsymbol{\omega}_{eb}^b \times \mathbf{l}^b \right) \right) + \mathbf{g}^n \tag{3}$$

because $\mathbf{l}^b$ is assumed to be constant. Organizing the terms and using (1) in [1] yield

$$\dot{\mathbf{v}}_{gps}^n + \left( 2\boldsymbol{\omega}_{ie}^n + \boldsymbol{\omega}_{en}^n \right) \times \mathbf{v}_{gps}^n - \mathbf{g}^n = \mathbf{C}_b^n \left[ \mathbf{f}^b + \dot{\boldsymbol{\omega}}_{eb}^b \times \mathbf{l}^b + \left( \boldsymbol{\omega}_{ie}^b + \boldsymbol{\omega}_{ib}^b \right) \times \left( \boldsymbol{\omega}_{eb}^b \times \mathbf{l}^b \right) \right]. \tag{4}$$

Making use of the chain rule of attitude matrix (cf. (4) and (5) in [1]), we have

---

This work was supported in part by the Fok Ying Tung Foundation (131061), National Natural Science Foundation of China (61174002), the Foundation for the Author of National Excellent Doctoral Dissertation of People's Republic of China (FANEDD 200897) and Program for New Century Excellent Talents in University (NCET-10-0900).

Authors' address: School of Aeronautics and Astronautics, Central South University, Changsha, Hunan 410083, P. R. China, Email: (yuanx_wu@hotmail.com).

$$\mathbf{C}_{n(t)}^{n(0)}\left[\dot{\mathbf{v}}_{gps}^{n}+\left(2\boldsymbol{\omega}_{ie}^{n}+\boldsymbol{\omega}_{en}^{n}\right)\times\mathbf{v}_{gps}^{n}-\mathbf{g}^{n}\right]=\mathbf{C}_{b}^{n}(0)\mathbf{C}_{b(t)}^{b(0)}\left[\mathbf{f}^{b}+\dot{\boldsymbol{\omega}}_{eb}^{b}\times\mathbf{l}^{b}+\left(\boldsymbol{\omega}_{ie}^{b}+\boldsymbol{\omega}_{ib}^{b}\right)\times\left(\boldsymbol{\omega}_{eb}^{b}\times\mathbf{l}^{b}\right)\right]. \quad (5)$$

Integrating (5) on both sides over the time interval of interest $[0,\ t]$

$$\int_{0}^{t}\mathbf{C}_{n(t)}^{n(0)}\dot{\mathbf{v}}_{gps}^{n}dt+\int_{0}^{t}\mathbf{C}_{n(t)}^{n(0)}\left(2\boldsymbol{\omega}_{ie}^{n}+\boldsymbol{\omega}_{en}^{n}\right)\times\mathbf{v}_{gps}^{n}dt-\int_{0}^{t}\mathbf{C}_{n(t)}^{n(0)}\mathbf{g}^{n}dt$$
$$=\mathbf{C}_{b}^{n}(0)\left[\int_{0}^{t}\mathbf{C}_{b(t)}^{b(0)}\mathbf{f}^{b}dt+\int_{0}^{t}\mathbf{C}_{b(t)}^{b(0)}\left(\dot{\boldsymbol{\omega}}_{eb}^{b}\times+\left(\left(\boldsymbol{\omega}_{ie}^{b}+\boldsymbol{\omega}_{ib}^{b}\right)\times\right)\left(\boldsymbol{\omega}_{eb}^{b}\times\right)\right)dt\,\mathbf{l}^{b}\right]. \quad (6)$$

The integral on the left $\int_{0}^{t}\mathbf{C}_{n(t)}^{n(0)}\dot{\mathbf{v}}_{gps}^{n}dt$ is developed as

$$\int_{0}^{t}\mathbf{C}_{n(t)}^{n(0)}\dot{\mathbf{v}}_{gps}^{n}dt=\mathbf{C}_{n(t)}^{n(0)}\mathbf{v}_{gps}^{n}-\mathbf{v}_{gps}^{n}(0)-\int_{0}^{t}\mathbf{C}_{n(t)}^{n(0)}\boldsymbol{\omega}_{in}^{n}\times\mathbf{v}_{gps}^{n}dt \quad (7)$$

where $\mathbf{v}_{gps}^{n}(0)$ is the initial velocity of the GPS antenna. The integral $\int_{0}^{t}\mathbf{C}_{b(t)}^{b(0)}\dot{\boldsymbol{\omega}}_{eb}^{b}\times dt$ is developed as

$$\int_{0}^{t}\mathbf{C}_{b(t)}^{b(0)}\dot{\boldsymbol{\omega}}_{eb}^{b}\times dt=\mathbf{C}_{b(t)}^{b(0)}\boldsymbol{\omega}_{eb}^{b}\times-\boldsymbol{\omega}_{eb}^{b}(0)\times-\int_{0}^{t}\mathbf{C}_{b(t)}^{b(0)}\left(\boldsymbol{\omega}_{ib}^{b}\times\right)\left(\boldsymbol{\omega}_{eb}^{b}\times\right)dt. \quad (8)$$

Substituting (7)-(8) into (6), we obtain

$$\mathbf{C}_{n(t)}^{n(0)}\mathbf{v}_{gps}^{n}-\mathbf{v}_{gps}^{n}(0)+\int_{0}^{t}\mathbf{C}_{n(t)}^{n(0)}\boldsymbol{\omega}_{ie}^{n}\times\mathbf{v}_{gps}^{n}dt-\int_{0}^{t}\mathbf{C}_{n(t)}^{n(0)}\mathbf{g}^{n}dt$$
$$=\mathbf{C}_{b}^{n}(0)\left[\int_{0}^{t}\mathbf{C}_{b(t)}^{b(0)}\mathbf{f}^{b}dt+\left(\mathbf{C}_{b(t)}^{b(0)}\boldsymbol{\omega}_{eb}^{b}\times-\boldsymbol{\omega}_{eb}^{b}(0)\times+\left(\boldsymbol{\omega}_{ie}^{b(0)}\times\right)\int_{0}^{t}\mathbf{C}_{b(t)}^{b(0)}\left(\boldsymbol{\omega}_{eb}^{b}\times\right)dt\right)\mathbf{l}^{b}\right] \quad (9)$$
$$\approx\mathbf{C}_{b}^{n}(0)\left[\int_{0}^{t}\mathbf{C}_{b(t)}^{b(0)}\mathbf{f}^{b}dt+\left(\mathbf{C}_{b(t)}^{b(0)}\boldsymbol{\omega}_{ib}^{b}\times-\boldsymbol{\omega}_{ib}^{b}(0)\times\right)\mathbf{l}^{b}\right].$$

The above equation is of the identical form to the velocity integration formula ((11)-(12) in [1]), except that the right side of (9) consists of an additional term related to the lever arm, namely, $\left(\mathbf{C}_{b(t)}^{b(0)}\boldsymbol{\omega}_{ib}^{b}\times-\boldsymbol{\omega}_{ib}^{b}(0)\times\right)\mathbf{l}^{b}$, a function of gyroscope/accelerometer outputs and the known lever arm. The approximation above is reasonable, since the Earth angular rate $\boldsymbol{\omega}_{ie}^{b}$ is negligible in practice with respect to the INS body-related angular rates $\boldsymbol{\omega}_{ib}^{b}$ or $\boldsymbol{\omega}_{eb}^{b}$.

*Remark 2.1*: It will be a good approximation if $\left\|\mathbf{C}_{b(t)}^{b(0)}\boldsymbol{\omega}_{eb}^{b}\times\right\|\gg\left\|\left(\boldsymbol{\omega}_{ie}^{b(0)}\times\right)\int_{0}^{t}\mathbf{C}_{b(t)}^{b(0)}\left(\boldsymbol{\omega}_{eb}^{b}\times\right)dt\right\|$ in magnitude. Considering the small magnitude of Earth rotation rate ($\sim 7.3\times 10^{-5}$ rad/s), the time duration for it being a valid approximation is no less than thousands of seconds.

*Remark 2.2*: The integral term $\int_{0}^{t}\mathbf{C}_{b(t)}^{b(0)}\mathbf{f}^{b}dt$ is generally increasing in magnitude as time goes, in contrast to the coefficient term related to the lever arm, so it can be inferred that the lever arm effect would be significantly mitigated after a while. It accords with our previous observations in [1] when the lever arm was not considered at all.

*Remark 2.3:* The position integration formula accounting for the prior knowledge of the lever arm can be similarly developed. It is straightforward and omitted here.

The simulation test in [1] is re-examined using (9) instead. Assume the GPS lever arm is precisely known beforehand by measurement or calibration, i.e., $\mathbf{l}^b = \begin{bmatrix} 1 & 1 & 1 \end{bmatrix}^T$ in meter. The mean alignment angle errors across 100 Monte Carlo runs after compensating the lever arm is presented in Fig. 1 and compared with those results in Fig. 5 in [1]. The lump peaks due to the presence of the lever arm are largely removed and the estimate errors are comparable to the simulated case with zero lever arm. It shows the remarkable effectiveness of (9) in depressing the lever arm effect.

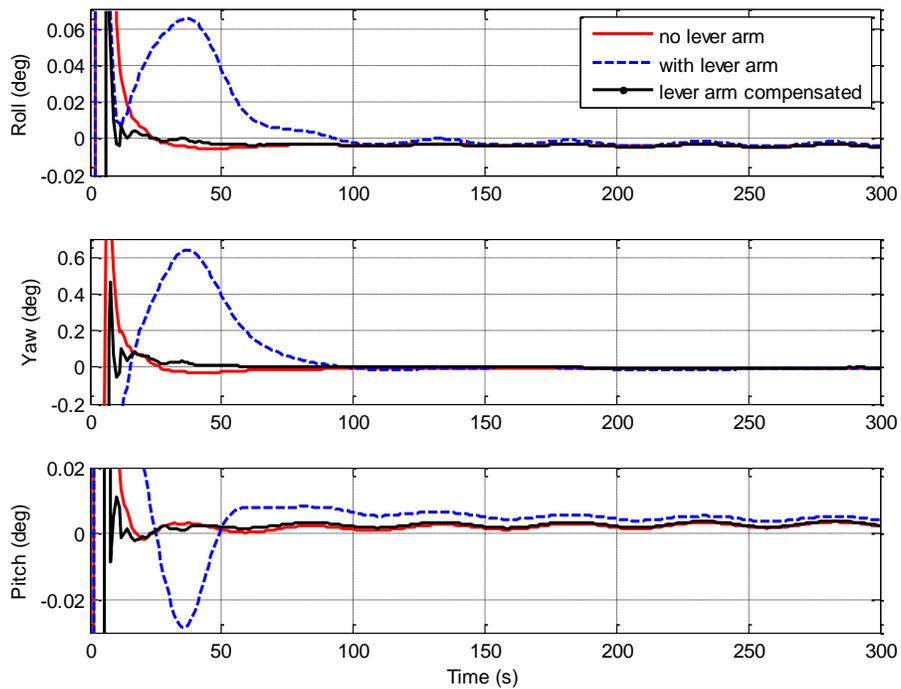

Figure 1. Mean alignment errors after compensating the lever arm, overlapped with those results in [1].

This note generalizes the algorism in the previous paper [1] by incorporating the prior knowledge of the lever arm. Results show that alignment performance is further enhanced in terms of accuracy and rapidness.

REFERENCES

[1]   Y. Wu and X. Pan, "Velocity/Position Integration Formula (I): Application to In-flight Coarse Alignment," *IEEE Trans. on Aerospace and Electronic Systems,* vol. 49, pp. 1006-1023, 2013.

[2]     P. D. Groves, *Principles of GNSS, Inertial, and Multisensor Integrated Navigation Systems*: Artech House, Boston and London, 2008.